\documentclass[runningheads]{llncs}

\input{psfig.sty}
\usepackage{algorithm,algpseudocode}
\usepackage{graphicx}
\usepackage{url}
\usepackage{xcolor}
\usepackage{etoolbox}
\patchcmd{\thebibliography}{\chapter*}{\section*}{}{}
\graphicspath{ {images/} }

\emergencystretch=\maxdimen
\begin{document}

\pagestyle{headings}

\mainmatter

 \title{An Image Dataset of\\
 Text Patches in Everyday Scenes}
 \titlerunning{Lecture Notes in Computer Science}
 \author{Ahmed Ibrahim\inst{1,4}, A. Lynn Abbott\inst{1}, Mohamed E. Hussein\inst{2,3} \\
 \institute{Virginia Polytechnic Institute and State University, USA\\
 \and Egypt-Japan University of Science and Technology, Egypt\\
 \and Alexandria University, Egypt\\
 \and Benha University, Egypt \\
\email{\{nady,abbott\}@vt.edu; mohamed.e.hussein@ejust.edu.eg}}
 }

\maketitle

\begin{abstract} 
    This paper describes a dataset containing  small images of text from everyday scenes. The purpose of the dataset is to support the development of new automated systems that can detect and analyze text. Although much research has been devoted to text detection and recognition in scanned documents, relatively little attention has been given to text detection in other types of images, such as photographs that are posted on social-media sites.  This new dataset, known as COCO-Text-Patch, contains approximately 354,000 small images that are each labeled as ``text'' or ``non-text''. 
This dataset particularly addresses the problem of text verification, which is an essential stage in the end-to-end text detection and recognition pipeline.
In order to evaluate the utility of this dataset, it has been used to train two deep convolution neural networks  to distinguish text from non-text. One network is inspired by the GoogLeNet  architecture, and the second one is based on CaffeNet. Accuracy levels of 90.2\% and 90.9\% were obtained using the two networks, respectively.  All of the images, source code, and deep-learning trained models described in this paper will be publicly available \footnote{\url{https://aicentral.github.io/coco-text-patch/}}.
\end{abstract}

\section{Introduction}

The ability to detect and recognize text in images of everyday scenes  will become increasingly important for such  applications as robotics and assisted driving. 
Most previous work involving text has focused on problems related to document analysis, or on optical character recognition (OCR) for text that has already been localized within images (e.g., reading automobile license plates). Unfortunately, text that appears in an unstructured setting can be difficult to locate and process automatically.

In spite of recent significant advances in automated analysis of text, everyday scenes continue to pose many challenges~\cite{2016scene_survey}.
For example, consider an image sequence taken from an automobile as it moves along a city street.
Text will be visible in a rich diversity of sizes, colors, fonts, and orientations. Furthermore, single lines of text may vary in scale due to perspective foreshortening.
Unlike most documents, the background may be very complex. Finally, other interference factors such as noise, motion blur, defocus blur, low resolution, nonuniform illumination, and partial occlusion may complicate the analysis.

\begin{figure}[h]
\centering
\includegraphics[width=\textwidth]{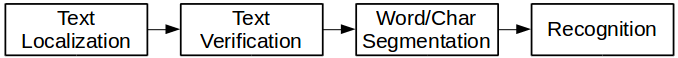}
\caption{The text detection and recognition pipeline. The 1st stage identifies candidate text locations in an image, and the 2nd stage removes false positives. The 3rd stage attempts to extract individual words or characters, which are recognized in the last stage.}
\label{fig:pipeline}
\end{figure}

Text detection and recognition can be divided into a pipeline of 4 stages as shown in Figure \ref{fig:pipeline}. The first stage, text localization, generates region proposals, which are typically rectangular sub-images that are likely to contain text. It is expected, however, that this stage will generate a relatively high number of false positives \cite{IEEE2016_survey}. The next stage, text verification, analyzes each region proposal further  in an attempt to remove false positives. 
The third stage, word/character segmentation, attempts to locate individual words or characters within the surviving region proposals.
The last stage, text recognition, is where OCR-type techniques are applied in an effort to recognize 
the extracted words and characters from the previous stage.

\begin{figure}[t!]
\centering
\includegraphics[width=0.8\textwidth,keepaspectratio]{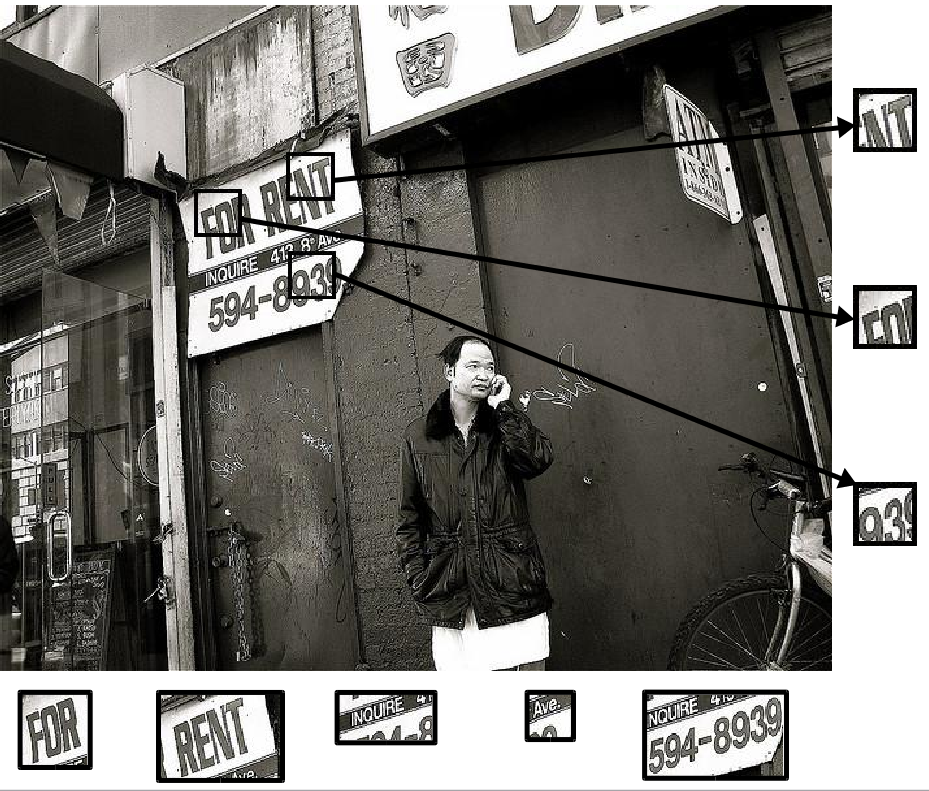}
\caption{Left: A sample image from COCO-Text \cite{cocotext}, 
with all text instances labeled as ``legible'' shown below it. Right: Several small images that are provided in the new dataset COCO-Text-Patch, which is introduced in this paper. Each small patch shown at the right is a $32 \times 32$ sample that contains text. The dataset also provides non-text (background) patches, as needed for training. }
\label{fig:cocotext}
\end{figure}

To aid in the development of automated text analysis systems, a number of image datasets have been collected and disseminated. 
One of these is COCO-Text~\cite{cocotext}, which contains 63,686 images of real-world scenes with  173,589 instances of text. 
An example from COCO-Text  is shown in Figure \ref{fig:cocotext}.
COCO-Text, which is described further in the next section, provides metadata to indicate the locations of text within the images.
These locations are given as rectangles that are aligned with the image borders. Limitations of COCO-Text are that character-level localization is not provided, nor are its text boxes in a format that is well suited for most deep-learning systems. 

The new dataset that is introduced here, COCO-Text-Patch, overcomes some of the limitations of COCO-Text by providing a large number of small images (``patches'')   that have been extracted from COCO-Text.
Each small image is labeled as ``text'' or ``non-text'', as needed for training,
with an emphasis on providing textural cues rather than instances of individual words or characters.
As shown in the examples of Figure \ref{fig:cocotext}, each small image in COCO-Text-Patch is of size $32 \times 32$ pixels, which is very well suited for many deep-learning implementations.
This new dataset will therefore satisfy an important need  in the development of new text-analysis systems.

Another consideration is that deep-learning systems require large numbers of training samples. As described in Section 3, COCO-Text-Patch provides small samples  in sufficient quantity and  in the proper format to support deep-learning methods.
The dataset has been balanced, so that approximately half of the patches contain text, and half represent background regions that do not contain text.

The rest of this paper is organized as follows. 
Section \ref{section:relatedwork} presents a brief survey of related work, including a discussion of datasets that have been collected to support research related to text detection.
Section \ref{datasetcreation} provides details concerning the new COCO-Text-Patch dataset.
In order to gauge the utility of this dataset, it was used in training two deep-learning networks, and these experiments are described in Section \ref{eval}. Finally, concluding remarks are given in Section \ref{section:conclusion}.

\section{Related Work}
\label{section:relatedwork}
\subsection{Text Verification}

The primary role of the text-verification  stage is to analyze tentative text regions from the text-localization stage and remove false positives~\cite{IEEE2016_survey}.
The text-verification  stage therefore plays an important role in enhancing overall system performance.
A variety of techniques have been utilized  for this task.
For example, Li and Wang used thresholds on the width and height of text blocks, along with other metrics such as the ratio between edge area and text area~\cite{txtverify1}.
Shivakumara et al. used profile projections, edge density, and character distances to filter out false positive regions~\cite{txtverify2,txtverify3,txtverify4}.  
Kim and Kim  used aspect ratios to verify the proposed text regions~\cite{txtverify5}.
Hanif et al. used height and width of connected components, edge counts, and horizontal profile projections~\cite{txtverify6}. 
Liu et al. used the ratio between width and height of minimum bounding rectangles, and the ratio between number of text and background pixels~\cite{txtverify8}.
Wang and Kangas are among the few who used OCR techniques to verify text regions~\cite{txtverify7}.
 
Recently, researchers have also begun to consider deep-learning approaches for text detection. 
With this strategy, the system learns features directly from actual pixel values in a set of training images.
In order to be effective, however, a large training set is needed. 
More details concerning this approach are given later in this section.

\subsection{Text Detection Datasets}
Many datasets have been created to help with various tasks related to text analysis. Example tasks include text detection, numerical digit  recognition, character recognition, and word-level text recognition. This section describes several popular datasets that have been used in text detection research.

Images in text-detection datasets can be grouped into two main categories. The first is 
\textit{focused scene text},  where text is a major emphasis of the image, and most of the text instances are near the center of the image. This category includes the majority of the datasets listed below. 
The second category is \textit{incidental text},  where text is not the emphasis of the image.
This category is much harder to analyze because text can be anywhere in the image, and sometimes in a very small portion of the image. 

One of the best known datasets is MNIST~\cite{MNIST}, which was released in 1998. The dataset contains samples of the handwritten digits 0 to 9 in monochromatic images of size $32 \times 32$ pixels. MNIST contains 60,000 samples  in a training set, and another 10,000 images in a test set. MNIST is widely used in tutorials because of its simplicity and relatively small size. 

The ICDAR 2003 dataset~\cite{ICDAR2003} was released as a part of an automated reading competition organized by the International Conference of Document Analysis and Recognition.    
This dataset contains 258 annotated images to be used for training, and 251 annotated images to be used for  validation. 
The images are in color, showing everyday scenes. Annotations provide rectangular bounding boxes to indicate instances of text.
This dataset was used for competitions in 2003 and 2005.

The ICDAR 2011 dataset~\cite{ICDAR2011} was created with an emphasis on finding text in born-digital images, particularly  images contained in web sites or in email communications. Born-digital images share most of the complexities associated with natural scene images. In addition, they often suffer from being lower in resolution. The dataset contains 420 images for training, and 102 images for testing. This dataset was used for automated reading competitions in 2011 and 2013.

The ICDAR 2015 robust reading competition introduced the first dataset devoted to incidental text~\cite{ICDAR2015}. This dataset contains 229 color images for training, and 233 images for testing. The competition also provided datasets for born-digital text and text in video.

The MSRA-RD500 dataset~\cite{MSRA}  was introduced by Yao et al. in 2012. An emphasis of this work was the detection of text at arbitrary orientations in natural images.
The dataset contains 500 everyday  indoor and outdoor images with English and Chinese text instances. 
Unlike previous datasets, 
MSRA-RD500 provides bounding boxes that have been rotated to accommodate instances of rotated text.
In other datasets, rotated text instances are simply annotated using larger rectangular boxes that are aligned with the image boundaries. 
Researchers who consider horizontal text only,  or who depend on  Latin language characteristics, tend to avoid this dataset.

The largest publicly available dataset to date that
supports text detection is COCO-Text~\cite{cocotext}. It was released early in 2016, and was derived from a larger dataset known as COCO (Common Objects in Context)~\cite{coco}. 
COCO was developed to support many computer-vision tasks, including image recognition, segmentation, and captioning. 
COCO-Text provides bounding rectangles along with a collection of labels for the text instances that appear in COCO images. 
COCO-Text contains a total of 63,686 images, with 43,686 training images and 20,000 validation images.
In addition to localization information, COCO-Text identifies text instances as machine-printed or handwritten, and legible or illegible.
Transcriptions are provided for the legible instances of text.
A limitation of COCO-Text is that it does not provide character level labeling, which could be used to  extract text sub-images. 
The COCO-Text-Patch dataset will help fill this gap. 

\subsection{Deep Learning}
Deep learning is a  machine-learning technique that has become increasingly popular in computer vision research. 
The main difference between classical  machine learning (ML) and deep learning is the way that features are extracted. 
For classical ML techniques such as support vector machines (SVM)~\cite{svm}, feature extraction is performed in advance using techniques crafted by the researchers. Then the training procedure develops weights or rules that map any given feature vector to an output class label. 
In contrast, the usual deep-learning procedure is to apply signal values as inputs directly to the ML network, without any preliminary efforts at feature extraction.
The network takes the input signal (pixel values, in our case), and assigns a class label based on those signal values directly. 
Because 
the deep-learning approach  implicitly must derive its own features, many more training samples are required than for traditional ML systems.

Several deep-learning packages are available for researchers. 
The package that we have used to evaluate COCO-Text-Patch is  Caffe \cite{caffe}, which is popular and was built with computer vision tasks in mind. Caffe is relatively easy to use, flexible, and powerful. It was developed using C++ code that utilizes GPU optimization libraries such as CuDNN, BLAS, and ATLAS. 

\section{Dataset Creation} \label{datasetcreation}

\subsection{Text and Non-text Patch Extraction}

\begin{figure}[b]
 \centering
 \includegraphics[width=0.99\textwidth,keepaspectratio]{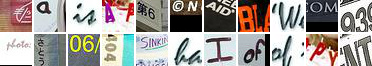}
 \caption{Sample text patches from COCO-Text-Patch. The text patches represent a wide range of visual textures, colors, font types, and character orientations. In order to emphasize textural cues over character shapes, no attempt was made to capture individual characters or words. }
 \label{fig:samples}
 \end{figure}

The procedure for extracting small text patches was relatively straightforward. 
For every legible text box that has been indicated by COCO-Text, including both machine-printed and handwritten cases, our system extracted non-overlapping sub-images of size $32 \times 32$  directly from the original COCO images.
The patch dimensions were chosen largely because this size is convenient 
for some of the popular deep network architectures. Most of the deep network architectures that are designed for small image datasets such as MNIST, CIFAR10, and CIFAR100 expect input images to be of size $32 \times 32$. The other widely used image size used in training deep networks is $256 \times 256$. This size is suitable for representing complex scenes with multiple instances of objects, which is not the case for text patches.
A few examples of the resulting COCO-Text-Patch images are shown in Figure \ref{fig:samples}.  

Similarly, small non-text patches were extracted from portions of COCO images that are outside the text boxes indicated by COCO-Text.
Text, by its nature, implies significant variations in visual texture. It was therefore important for COCO-Text-Patch to provide  non-text examples that contain substantial levels of texture. For this reason, a texture-based measure was employed during the balancing step, as described in the next section.
Python was used to implement all extraction and balancing algorithms. 

\subsection{Dataset Balancing}

\begin{figure}[h]
  \centering
 \includegraphics[width=0.69\textwidth,height=\textheight,keepaspectratio]{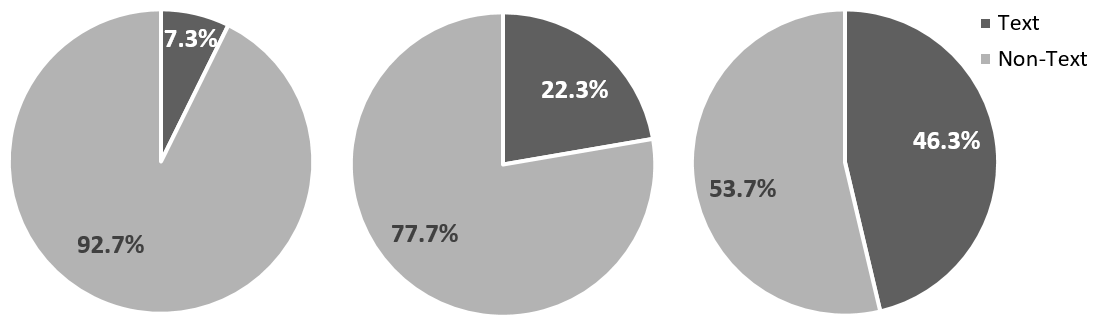}
 \caption{The ratio between text and non-text instances during development of COCO-Text-Patch. Left: the initial proportion of legible machine-printed text patches to non-text patches was approximately  7\% to 93\%. Center: increased proportion after inclusion of legible handwritten text.  
Right: final well-balanced proportion, after texture-based filtering of non-text patches.}
 \label{fig:balance}
 \end{figure}

Because text represents a relatively small proportion of image area within  COCO-Text images, many more non-text patches than text patches were detected initially using the extraction strategy described in the previous section.
In fact, as indicated in Figure \ref{fig:balance}, the number of patches extracted from legible machine-printed text represented  less than 10\% of the patches that were initially extracted.
When text patches were also extracted from legible handwritten text, the proportion of text patches rose to about 22\%.

In order to support ML approaches, particularly deep-learning, it was decided to provide further balance to COCO-Text-Patch by removing some of the non-text cases.
Random sampling was considered briefly.
However, the importance of texture led us to implement a fast texture-based approach.
In our implementation, this analysis was accomplished by applying  Prewitt \cite{prewitt} filters to
a grayscale version of each non-text patch. 
The resulting gradient magnitudes were binarized, to indicate the presence of intensity edges.
If the number of edge pixels exceeded an empirically selected threshold, then the patch was retained as a non-text sample.
This edge-count threshold was adjusted so that a split of approximately 50:50 for text:non-text was achieved. 
As shown in the figure, the actual final ratio in the COCO-Text-Patch dataset was close to 46:54.
Actual image quantities are shown in  Table \ref{cocopatchtable}. 

\begin{table}[h]
\centering
\caption{Number of patches in the final COCO-Text-Patch dataset.}
\label{cocopatchtable}
\begin{tabular}{|c|c|c|c|}
\hline
    & ~Text~     & ~Non-text~    & ~Total~  \\ \hline
~Training~   & ~112044~   & ~130041~   & ~242085~ \\ \hline
~Validation~ & ~52085~   & ~59977~   & ~112062 \\ \hline
~Total~     & ~164129~ & ~190018~ & ~354147~ \\ \hline

\end{tabular}
\end{table}

\section{Evaluation}
\label{eval}

The utility of the new COCO-Text-Patch dataset was evaluated using 
two convolutional neural networks.
Both  CaffeNet \cite{caffenet} and GoogLeNet \cite{googlenet} were trained using Caffe \cite{caffe}, and the resulting models will be made available to the research community. 
All training and testing has been done on the Virginia Tech  NewRiver HPC~\cite{newriver}.

\subsection{Models}
Both CaffeNet and GoogLeNet were created to be used with images of size $256 \times $256, and each was designed to learn 1000 classes. 
We modified both network architectures  to be able to learn from the smaller size $32 \times $32 images in COCO-Text-Patch as well as being able to learn 2 classes instead of 1000. The modified CaffeNet and GoogLeNet will be referred to as CaffeNet-TXT and GoogLeNet-TXT, respectively. 

A close examination of the layers of CaffeNet  show that the network tries to perform dimension reduction in the early layers. This can be inferred from the parameters of the CONV1  and POOL1 layers. The stride of CONV1 is set to 4, which was originally chosen to reduce the input size by a factor of four from $256 \times $256 to $64 \times $64.
Then the POOL1 layer with stride 3 would cause more reduction. Those rapid reductions are not suitable for a input size of $32 \times $32, so those layers have been modified such that  the CONV1 and POOL1 layers both have a stride of 1. Similar stride reduction has been performed on the GoogLeNet architecture as well.

The CaffeNet and GoogLeNet architectures can both  be  viewed simply as a set of convolutional layers followed by a set of fully connected layers, which is  then followed by a softmax layer that will generate a one-hot class label output. Both networks are typically set to have 1000 outputs in the last fully connected layer, corresponding to 1000 classes. The last fully connected layer of both networks has been modified to have 2 outputs corresponding to the  classes text and non-text.

\subsection{Experiments and Results}
We conducted 4 experiments  using the COCO-Text-Patch dataset. Two experiments used the CaffeNet-TXT architecture.
The first experiment used a preliminary dataset that was balanced using random sampling, and the second experiment used the final dataset that was  balanced using texture analysis. The other two experiments used GoogLeNet-TXT, with the same two datasets.

The best average accuracy for the dataset that was balanced using random sampling was 90.1\% (using GoogLeNet), while the best average accuracy for the dataset that was balanced using texture analysis was 90.9\% (using CaffeNet). The  results are summarized in Table \ref{table_results}.

\begin{table}[]
\centering
\caption{Evaluation results for the new COCO-Text-Patch dataset. Two methods of balancing were performed: textural analysis (left) and random selection (right), with the former yielding better accuracy values. Two architectures were considered: CaffeNet and GoogLeNet. The lower part of the table contains a confusion matrix for each of the four experiments.}
\label{table_results}
\begin{tabular}{|c|c|c|c|c|c|c|c|c|}
\hline
~Balancing~ & \multicolumn{4}{c|}{Texture Analysis} & \multicolumn{4}{c|}{Random Sampling}  \\ \hline
Network & \multicolumn{2}{c|}{CaffeNet} & \multicolumn{2}{c|}{GoogLeNet} & \multicolumn{2}{c|}{CaffeNet} & \multicolumn{2}{c|}{GoogLeNet} \\ \hline
Accuracy & \multicolumn{2}{c|}{90.9\%} & \multicolumn{2}{c|}{90.2\%} & \multicolumn{2}{c|}{85.2\%} & \multicolumn{2}{c|}{90.1\%} \\ \hline
       &   Text     & Non-text  & Text     & Non-text &   Text     & Non-text  & Text     & Non-text     \\ \hline
Text   &  0.868   & 0.057   & 0.838 & 0.042 &   0.856   & 0.054   & 0.729 & 0.022 \\ \hline
Non-text & 0.132   & 0.943   & 0.162 &  0.958 & 0.144   &  0.946   & 0.271 &  0.978 \\ \hline

\end{tabular}
\end{table}

The lower accuracy that was obtained using random sampling
may be due to the lower  proportion of non-text training samples having significant levels of texture.
If a substantial number of low-texture training samples are used, then the final system may be biased in a way that favors low texture in order to receive the non-text label.
This bias is reduced somewhat for the case that texture-based selection was used for balancing.

\section{Conclusion}
\label{section:conclusion}
This paper has introduced the COCO-Text-Patch dataset, which will support the development of automated  systems that detect and analyze text in natural and everyday images.
The emphasis has been to provide a
dataset that can be used as a standard for evaluating the text verification step.  
The COCO-Text-Patch  training and validation sets have been created from the COCO-Text training and validation sets respectively. A texture-based approach was used to balance the dataset, so that it contains nearly equal numbers of text and non-text samples.
Evaluation has been done using 2 deep neural networks, and an accuracy of 90.9\% was observed for CaffeNet.
Such a result suggests that future text-analysis systems can benefit substantially through training that includes the COCO-Text-Patch dataset.
{\footnotesize
\bibliography{isvc_submission}
}

\end{document}